\documentclass[11pt,a4paper]{article}
\usepackage[hyperref]{acl2019}
\usepackage{times}
\usepackage{latexsym}
\usepackage{url}  %Required
\usepackage{graphicx}  %Required
\usepackage{color}
\usepackage{amsmath}
\usepackage{amssymb}
\usepackage{algorithm}
\usepackage{setspace,lipsum}
\usepackage[noend]{algpseudocode}
\usepackage{booktabs}
\aclfinalcopy % Uncomment this line for the final submission
\def\aclpaperid{1465} %  Enter the acl Paper ID here

\title{Unsupervised Neural Text Simplification}
\author{Sai Surya$^{\dagger}$ \hspace{0.05cm} Abhijit Mishra$^{\ddagger}$ \hspace{0.05cm} Anirban Laha$^{\ddagger}$ \hspace{0.05cm} \bf{Parag Jain$^{\ddagger}$ \hspace{0.05cm} Karthik Sankaranarayanan}$^{\ddagger}$\\
  $^\dagger$IIT Kharagpur, India \hspace{0.1cm} 
  $^{\ddagger}$IBM Research\\
  {\tt subramanyamdvss@gmail.com} \\ {\tt \{abhijimi,anirlaha,pajain34,kartsank\}@in.ibm.com}
}

\date{}
\begin{document}
\maketitle
\begin{abstract}
The paper presents a first attempt towards unsupervised neural text simplification that relies only on unlabeled text corpora. The core framework is composed of a shared encoder and a pair of attentional-decoders, crucially assisted by \emph{discrimination-based losses} and \emph{denoising}. The framework is trained using unlabeled text collected from \textit{en-Wikipedia} dump. Our analysis (both quantitative and qualitative involving human evaluators) on public test data shows that the proposed model can perform text-simplification at both lexical and syntactic levels, competitive to existing supervised methods. It also outperforms viable unsupervised baselines. Adding a few labeled pairs helps improve the performance further. %We will open source our implementation for academic research.
\end{abstract}

%%%%%%%%%%%Sectionwise input %%%%%%%%%%%
\section{Introduction}
\label{sec:intro}
Text Simplification (TS) deals with transforming the original text into simplified variants to increase its readability and understandability. TS is an important task in computational linguistics, and has numerous use-cases in fields of education technology, targeted content creation, language learning, where producing variants of the text with varying degree of simplicity is desired. TS systems are typically designed  to simplify from two different linguistic aspects: (a) \textbf{Lexical} aspect, by replacing complex words in the input with simpler synonyms \cite{devlin1998use,candido2009supporting,yatskar2010sake,biran2011putting,glavavs2015simplifying}, and  (b) \textbf{Syntactic} aspect, by altering the inherent hierarchical structure of the sentences \cite{chandrasekar1997automatic,canning1999syntactic,siddharthan2006syntactic,filippova2008dependency,brouwers2014syntactic}. From the perspective of sentence construction, sentence simplification can be thought to be a form of  \textit{text-transformation} that involves three major types of operations such as (a) \textit{splitting} \cite{siddharthan2006syntactic,petersen2007text,narayan2014hybrid} (b) \textit{deletion/compression} \cite{knight2002summarization,clarke2006models,filippova2008dependency,rush2015neural,filippova2015sentence}, and (c) \textit{paraphrasing}  \cite{specia2010translating,coster2011simple,wubben2012sentence,wang2016text,nisioi2017exploring}.

Most of the current TS systems require large-scale parallel corpora for training (except for systems like \newcite{glavavs2015simplifying} that  performs only lexical-simplification), which is a major impediment in scaling to newer languages, use-cases, domains and output styles for which such large-scale parallel data do not exist. In fact, one of the popular corpus for TS in English language, \textit{i.e.,} the Wikipedia-SimpleWikipedia aligned dataset has been prone to noise (mis-aligned instances) and inadequacy (i.e., instances having non-simplified targets) \cite{xu2015problems,vstajner2015deeper}, leading to noisy supervised models \cite{wubben2012sentence}. While creation of better datasets (such as, \textit{Newsela} by \newcite{xu2015problems}) can always help, we explore the unsupervised learning paradigm which can potentially work with unlabeled datasets that are cheaper and easier to obtain. 

At the heart of the TS problem is the need for preservation of language semantics with the goal of improving readability. From a neural-learning perspective, this entails a specially designed \emph{auto-encoder}, which not only is capable of reconstructing the original input but also can additionally introduce variations so that the auto-encoded output is a simplified version of the input. Intuitively, both of these can be learned by looking at the structure and language patterns of a large amount of non-aligned complex and simple sentences (which are much cheaper to obtain compared to aligned parallel data). These motivations form the basis of our work.

Our approach relies only on two unlabeled text corpora - one representing relatively simpler sentences than the other (which we call complex). The crux of the (\emph{\textbf{unsupervised}}) auto-encoding framework is a shared encoder and a pair of attention-based decoders (one for each type of corpus). The encoder attempts to produce semantics-preserving representations which can be acted upon by the respective decoders (simple or complex) to generate the appropriate text output they are designed for. The framework is crucially supported by two kinds of losses: (1) \emph{\textbf{adversarial loss}} - to distinguish between the real or fake attention context vectors for the simple decoder, and (2) \emph{\textbf{diversification loss}} - to distinguish between attention context vectors of the simple decoder and the complex decoder. The first loss ensures that only the aspects of semantics that are necessary for simplification are passed to the simple decoder in the form of the attention context vectors. The second loss, on the other hand, facilitates passing different semantic aspects to the different decoders through their respective context vectors. Also we employ \emph{\textbf{denoising}} in the auto-encoding setup for enabling syntactic transformations.

The framework is trained using unlabeled text collected from Wikipedia (complex) and Simple Wikipedia (simple). It attempts to perform simplification both lexically and syntactically unlike prevalent systems which mostly target them separately. We demonstrate the competitiveness of our unsupervised framework alongside supervised skylines through both automatic evaluation metrics and human evaluation studies. We also outperform another unsupervised baseline \cite{artetxe2018iclr}, first proposed for neural machine translation. Further, we demonstrate that by leveraging a small amount of labeled parallel data, performance can be improved further. Our code and a new dataset containing partitioned unlabeled sets of simple and complex sentences is publicly available\footnote{https://github.com/subramanyamdvss/UnsupNTS}. 
\section{Related Work}
\label{sec:related}
% ,crossley2007linguistic
% Text simplification, often discussed from psychological and linguistic standpoints \cite{lallier80eval,mcnamara1996good,linderholm2000effects}, recently became a mainstream research in computational linguistics. 
Text Simplification has often been discussed from psychological and linguistic standpoints \cite{lallier80eval,mcnamara1996good,linderholm2000effects}. A heuristic-based system was first introduced by \newcite{chandrasekar1997automatic} which induces rules for simplification automatically extracted from annotated corpora. \newcite{canning1999syntactic} proposed a modular system that uses NLP tools such as morphological analyzer, POS tagger plus heuristics to simplify the text both lexically and syntactically. Most of these systems \cite{siddharthan2014survey} are separately targeted towards lexical and syntactic simplification and are limited to splitting and/or truncating sentences.   
% With the advent of data-driven machine translation paradigms, it became possible to perform paraphrasing based simplification. Some of the initial systems, based on the Statistical Machine Translation (SMT), rely on
For paraphrasing based simplification, data-driven approaches were proposed like phrase-based SMT \cite{specia2010translating,vstajner2015deeper} or their variants \cite{coster2011simple,xu2016optimizing}, that combine heuristic and optimization strategies for better TS. Recently proposed TS systems are based on neural \emph{seq2seq} architecture \cite{bahdanau2014neural} which is modified for TS specific operations \cite{wang2016text,nisioi2017exploring}. While these systems produce state of the art results on the popular Wikipedia dataset \cite{coster2011simple}, they may not be generalizable because of the noise and bias in the dataset \cite{xu2015problems} and overfitting. %of supervised models on small scale labeled datasets. 
Towards this, \newcite{stajner2018} showed that improved datasets and minor model changes (such as using reduced vocabulary and enabling copy mechanism) help obtain reasonable performance for both in-domain and cross-domain TS. 

In the unsupervised paradigm, \newcite{paetzold2016unsupervised} proposed an unsupervised lexical simplification technique that replaces complex words in the input with simpler synonyms, which are extracted and disambiguated using word embeddings. However, this work, unlike ours only addresses lexical simplification and cannot be trivially extended for other forms of simplification such as splitting and rephrasing. Other works related to style transfer \cite{zhang2018style,shen2017style,xu2018unpaired} typically look into the problem of sentiment transformation and are not motivated by the linguistic aspects of TS, and hence not comparable to our work. As far as we know, ours is a first of its kind end-to-end solution for unsupervised TS. At this point, though supervised solutions perform better than unsupervised ones, we believe unsupervised techniques should be further explored since they hold greater potential with regards to scalability to various tasks.
% \textcolor{red}{Add next para}

% It is worth noting that in the absence of other end-to-end unsupervised baselines for TS, we train the unsupervised machine translation system by   \newcite{artetxe2017unsupervised} and consider it as a baseline.
\section{Model Description}
\label{sec:system}
%%%%%%%%%%%%%%%%Main Figure%%%%%%%%%%%%
\begin{figure*}[t]
\centering 
\includegraphics[width=10cm,height=10.345cm,keepaspectratio]{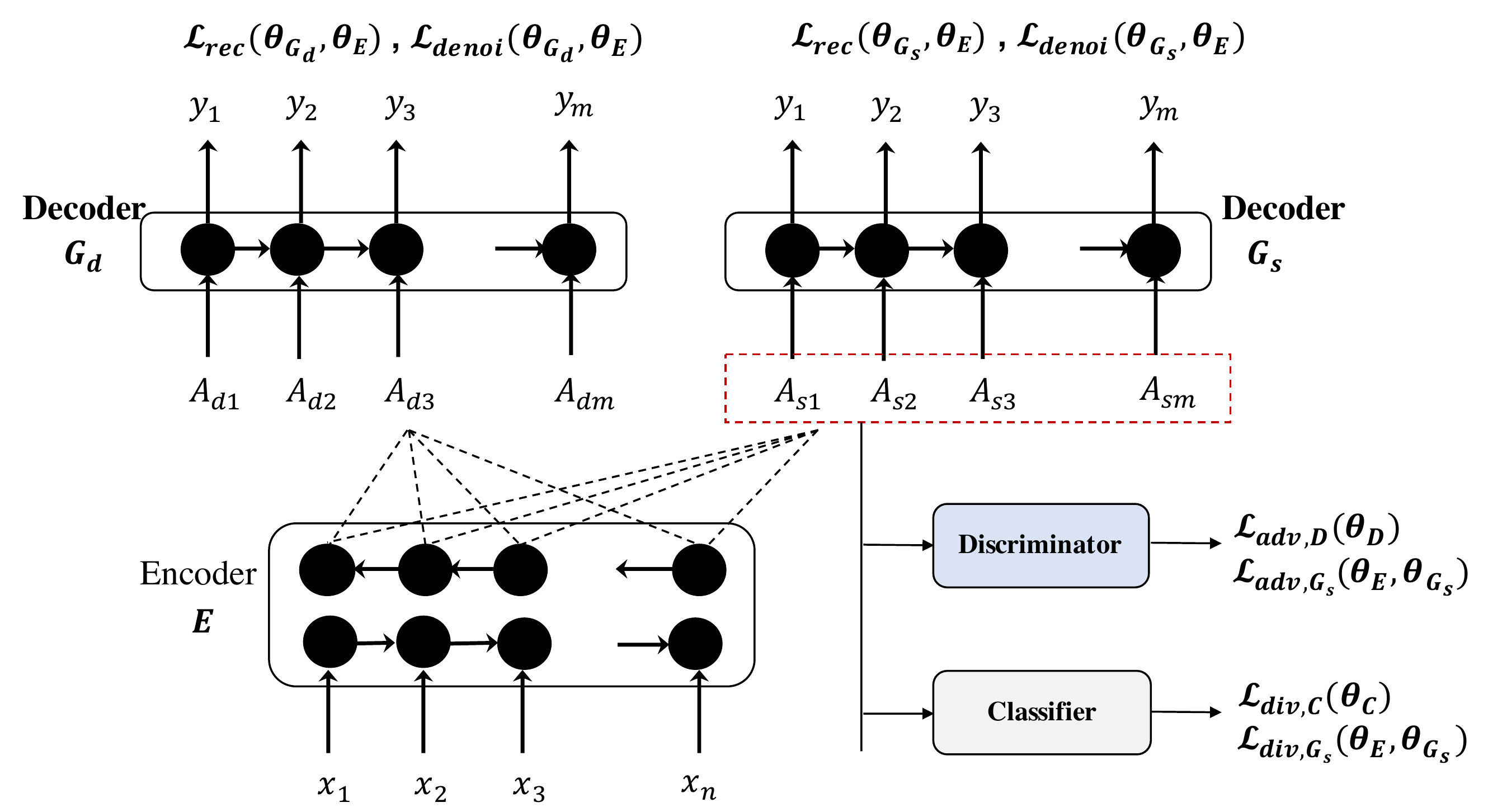}
\caption{System Architecture. Input sentences of any domain is encoded by $\boldsymbol{E}$, and decoded by $\boldsymbol{G_s}$, $\boldsymbol{G_d}$. Discriminator $\boldsymbol{D}$ and classifier $\boldsymbol{C}$ tune the attention vectors for simplification. $\mathcal{L}$ represents loss functions. The figure only reveals one layer in $\boldsymbol{E}$, $\boldsymbol{G_s}$ and $\boldsymbol{G_d}$ for simplicity. However, the model uses two layers of GRUs (Section \ref{sec:system}).}
\label{fig:workflow}
\end{figure*}
%%%%%%%%%%%%%%%%Main Figure%%%%%%%%%%%%
Our system is built based on the encode-attend-decode style architecture \cite{bahdanau2014neural} with both algorithmic and architectural changes applied to the standard model. An input sequence of word embeddings $X=\{x_1,x_2,\ldots,x_n\}$ (obtained after a standard look up operation on the embedding matrix), is passed through a shared encoder ($\boldsymbol{E}$), the output representation from which is fed to two decoders ($\boldsymbol{G_s}$, $\boldsymbol{G_d}$) with attention mechanism. $\boldsymbol{G_s}$ is meant to generate a simple sentence from the encoded representation, whereas $\boldsymbol{G_d}$ generates a complex sentence. A discriminator ($\boldsymbol{D}$) and a classifier ($\boldsymbol{C}$) are also employed adversarially to distinguish between the attention context vectors computed with respect to the two decoders. Figure \ref{fig:workflow} is illustrates our system. We describe the components below. 

\subsection{Encode-Attend-Decode Model}
\label{sec:enc-dec}
Encoder $\boldsymbol{E}$ uses two layers of bi-directional GRUs \cite{cho2014learning}, and decoders $\boldsymbol{G_s}$, $\boldsymbol{G_d}$ have two layers of GRUs each. $\boldsymbol{E}$ extracts the hidden representations from an input sentence. The decoders output sentences, sequentially one word at a time. Each decoder-step involves using \textit{global attention} to create a context-vector (hidden representations weighted by attention weights) as an input for the next decoder-step. The attention mechanism enables the decoders to focus on different parts of the input sentence. For the input sentence $X$ with $n$ words, the encoder produces $n$ hidden representations, $H=\{h_1,h_2,\ldots,h_n\}$. The context vector extracted from $X$ by a decoder $\boldsymbol{G}$ for time-step $t$ is represented as, 
\begin{equation}
A_t(X) = \sum_{i=1}^n a_{it}h_i
\label{eqn:attention}
\end{equation}
where, $a_{it}$ denotes attention weight for the hidden representation at the $i^{th}$ input position with respect to decoder-step $t$.
As there are two decoders, $A_{st}(X)$ and $A_{dt}(X)$ denote the context vectors computed from decoders $\boldsymbol{G_s}$ and $\boldsymbol{G_d}$ respectively for time-steps $t\in\{1\ldots m\}$, $m$ denoting the total number of decoding steps performed\footnote{For a particular $X$, $m$ can differ for the two decoders.}. The matrices $A_s(X)$ and $A_d(X)$ represent the sequence of respective context vectors from all time-steps.
% denote the hidden representations extracted by encoder $E$ from $X$, $A_t$ denotes attention vector extracted by the decoder at step $t$, $\{a_{1t},a_{2t},\ldots,a_{nt}\}$ denote the attention weights for the hidden representations.
% \begin{align}
% \label{eqn:attention}
% \begin{gathered}
% h_1,h_2,\ldots,h_n = \boldsymbol{E}({x_1,x_2,\ldots,x_n})\\ 
% A({x_1,x_2,\ldots,x_n}) = \{\sum_{i=1}^n a_{it}h_i\}_{t=1}^m
% \end{gathered}
% \end{align}

\subsection{Discriminator and Classifier}
A discriminator $\boldsymbol{D}$ is employed to influence the way the decoder $\boldsymbol{G_s}$ will attend to the hidden representations, which has to be different for different types of inputs to the shared encoder $\boldsymbol{E}$ (simple vs complex). The input to $\boldsymbol{D}$ is the context vector sequence matrix $A_{s}$ pertaining to $\boldsymbol{G_s}$, and it produces a binary output, $\{1,0\}$, $1$ indicating the fact that the context vector sequence is close to a typical context vector sequence extracted from simple sentences seen in the dataset. $\boldsymbol{G_s}$ and $\boldsymbol{D}$ are indulged in an adversarial interplay through an adversarial loss function (see Section \ref{sec:advloss}), analogous to GANs \cite{goodfellow2014generative}, where the generator and discriminators, converge to a point where the distribution of the generations eventually resembles the distribution of the genuine samples. In our case, adversarial loss tunes the context vector sequence from a complex sentence by $\boldsymbol{G_s}$ to ultimately resemble the context vector sequence of simple sentences in the corpora. \emph{This ensures that the resultant context vector for $\boldsymbol{G_s}$ captures only the necessary language signals to decode a simple sentence}.

A classifier ($\boldsymbol{C}$) is introduced for diversification to ensure that the way decoder $\boldsymbol{G_s}$ attends to the hidden representations remains different from $\boldsymbol{G_d}$. It helps distinguish between simple and complex context vector sequences with respect to  $\boldsymbol{G_s}$ and $\boldsymbol{G_d}$ respectively. \emph{The classifier  diversifies the context vectors given as input to the different decoders. Intuitively, different linguistic signals are needed to decode a complex sentence \textit{vis-\'a-vis} a simple one}. Refer Section \ref{sec:clfloss} for more details. 

Both  $\boldsymbol{D}$ and $\boldsymbol{C}$ use a CNN-based classifier analogous to \newcite{kim2014convolutional}. All layers are shared between $\boldsymbol{D}$ and $\boldsymbol{C}$ except the fully-connected layer preceeding the softmax function.

\subsection{Special Purpose Word-Embeddings}
Pre-trained word embeddings are often seen to have positive impact on  sequence-to-sequence frameworks \cite{cho2014properties,qi2018and}. However, traditional embeddings are not good at capturing relations like synonymy \cite{tissier2017dict2vec}, which are essential for simplification. For this, our word-embeddings are trained using the \textit{Dict2Vec} framework\footnote{https://github.com/tca19/dict2vec}. \textit{Dict2Vec} fine-tunes the embeddings through the help of an external lexicon containing weak and strong synonymy relations. The system is trained on our whole unlabeled datasets and with seed synonymy dictionaries provided by \newcite{tissier2017dict2vec}. Our encoder and decoders share the same word embeddings. Moreover, the embeddings at the input side are kept static but the decoder embeddings are updated as training progresses. Details about hyper-parameters are given in Section \ref{subsec:param}.

\section{Training Procedure}
\label{sec:training}

\def\xs {$\boldsymbol{X_s}$}
\def\xd {$\boldsymbol{X_d}$}
\def\S {$\boldsymbol{\mathcal{S}}$}
\def\Sh {$\boldsymbol{\mathcal{S}_h}$}
\def\D {$\boldsymbol{\mathcal{D}}$} 
\def\Dh {$\boldsymbol{\mathcal{D}_h}$}
\def\belongsto {$\in$}
\def\Enc {$\boldsymbol{E}$}
\def\G {$\boldsymbol{G}$}
\def\Ds {$\boldsymbol{D_s}$}
\def\Dd {$\boldsymbol{D_d}$}
\def\C {$\boldsymbol{C}$}
\def\thetaE{\boldsymbol{\theta_E}}
\def\thetaG{\boldsymbol{\theta_G}}
\def\thetaGs{\boldsymbol{\theta_{G_s}}}
\def\thetaGd{\boldsymbol{\theta_{G_d}}}
\def\thetaDs{\boldsymbol{\theta_{D_s}}}
\def\thetaC{\boldsymbol{\theta_{C}}}
\def\thetaD{\boldsymbol{\theta_{D}}}
\def\thetaDd{\boldsymbol{\theta_{D_d}}}
\def\simplf{$\boldsymbol{E-G_s}$}
\def\simplfm{\boldsymbol{E-G_s}}
\def\pipe{\boldsymbol{E-G}}
\def\fpipe{\boldsymbol{E-G}}
\def\compl{$\boldsymbol{E-G_d}$}
\def\complm{\boldsymbol{E-G_d}}
\def\Lrec{\mathcal{L_rec}}

Let $\boldsymbol{\mathcal{S}}$ and $\boldsymbol{\mathcal{D}}$ be sets of simple and complex sentences respectively from large scale  unlabeled repositories of simple and complex sentences. Let $X_s$ denote a sentence sampled from the set of simple sentences $\boldsymbol{\mathcal{S}}$ and $X_d$ be a sentence sampled from the set of complex sentences $\boldsymbol{\mathcal{D}}$.
% Let $\boldsymbol{\mathcal{S}_a}$ and $\boldsymbol{\mathcal{D}_a}$ denote the set of simple and difficult attention vectors. $\boldsymbol{\mathcal{S}_a}$ and $\boldsymbol{\mathcal{D}_a}$ are extracted by decoders $\boldsymbol{G_s}$ and $\boldsymbol{G_d}$ respectively from $\boldsymbol{\mathcal{S}}$ and $\boldsymbol{\mathcal{D}}$ domains.
Let $\thetaE\ $ denote the parameters of $\boldsymbol{E}$ and $\thetaGs\ , \thetaGd\ $ denote the parameters of $\boldsymbol{G_s}$ and $\boldsymbol{G_d}$ respectively. Also, $\thetaC\ $ and $\thetaD\ $ are the parameters of the discriminator and the classifier modules.
% Let $A_S$ denote an attention vector sequence sampled from set $\boldsymbol{\mathcal{S}_a}$ and $A_D$ be an attention vector sequence sampled from the set $\boldsymbol{\mathcal{D}_a}$. 
Training the model involves optimization of the above parameters with respect to the following losses and \textit{denoising}, which are explained below.

\subsection{Reconstruction Loss}
\label{sec:reconst}
Reconstruction Loss is imposed on both \simplf\ and \compl\ paths. \simplf\ is trained to reconstruct sentences from \S\ and \compl\ is trained to reconstruct sentences from \D. 
Let $P_{\boldsymbol{E-G_s}}({X})$ and $P_{\boldsymbol{E-G_d}}(X)$ denote the reconstruction probabilities of an input sentence $X$ estimated by the \simplf\ and \compl\ models respectively. Reconstruction loss for \simplf\ and \compl\ , denoted by $\mathcal{L}_{rec}$ is computed as follows.
% \vspace{-5pt}
\begin{align}
\small
 \mathcal{L}_{rec}(\thetaE,\thetaGs,\thetaGd) = -\mathbb{E}_{X_s\boldsymbol{ \sim  \mathcal{S}}}[\log{P_{\simplfm}(X_s)}] - \nonumber \\
\small
 \mathbb{E}_{X_d\boldsymbol{ \sim  \mathcal{D}}}[\log{P_{\complm}(X_d)}]
\label{eqn:reconst}
\end{align}
\subsection{Adversarial Loss} 
\label{sec:advloss}
Adversarial Loss is imposed upon the context vectors for $\boldsymbol{G_s}$. The idea is that, context vectors extracted even for a complex input sentence by $\boldsymbol{G_s}$ should resemble the context vectors from a simple input sentence. The discriminator $\boldsymbol{D}$ is trained to distinguish the \textit{fake} (complex) context vectors from the \textit{real} (simple) context vectors. \simplf\ is trained to perplex the discriminator $\boldsymbol{D}$, and eventually, at convergence, learns to produce real-like (simple) context vectors from complex input sentences. In practice, we observe that adversarial loss indeed assists \simplf\ in simplification by encouraging sentence shortening. Let $A_s(.)$ be a sequence of context vectors as defined in Section \ref{sec:enc-dec}. Adversarial losses for \simplf\ , denoted by $\mathcal{L}_{adv,G_s}$ and for discriminator $\boldsymbol{D}$, denoted by $\mathcal{L}_{adv,D}$ are as follows.
% Let hidden state sequence $\boldsymbol{h_s}$ \belongsto \Sh\ and $\boldsymbol{h_d}$ \belongsto \Dh. 
% Our model is based on assumption that while transferring from \D\ to \S\ the attended vectors $A_{\boldsymbol{G_s}}(\boldsymbol{h_d})$ generated should be having same distribution as $A_{\boldsymbol{G_s}}(\boldsymbol{h_s})$. This reduces the content extracted from $\boldsymbol{h_d}$ thus helping in simplification. So we employ adversarial losses $\mathcal{L}_{adv,G_s}$ for the decoder $\boldsymbol{G_s}$ and $\mathcal{L}_{adv,D}$ for discriminator.
% \vspace{-5pt}
\begin{align}
\small
\mathcal{L}_{adv,D}(\boldsymbol{\theta_{D}}) = -\mathbb{E}_{X_s\boldsymbol{ \sim  \mathcal{S}}}[\log{(\boldsymbol{D}(A_s (X_s)))}] - \nonumber\\ 
\small
\mathbb{E}_{X_d\boldsymbol{ \sim  \mathcal{D}}}[\log{(1- \boldsymbol{D}(A_s (X_d))}] \\
\small
\mathcal{L}_{adv,G_s}(\thetaE,\thetaGs) = -\mathbb{E}_{X_d\boldsymbol{ \sim  \mathcal{D}}}[\log{(\boldsymbol{D}(A_s(X_d)))}]
\label{eqn:advloss}
\end{align}
% \vspace{-0.5in}
\subsection{Diversification Loss}
\label{sec:clfloss}
Diversification Loss is imposed by the classifier $\boldsymbol{C}$ on context vectors extracted by $\boldsymbol{G_d}$ from complex input sentences in contrast with context vectors extracted by $\boldsymbol{G_s}$ from simple input sentences. This helps \simplf\ to learn to generate simple context vectors distinguishable from complex context vectors. Let $A_s(.)$ and $A_d(.)$ be sequence of context vectors as defined in Section \ref{sec:enc-dec}. Losses for classifier C, denoted by $\mathcal{L}_{div,C}$ and for model \simplf\, denoted by $\mathcal{L}_{div,G_s}$ are computed as follows.
% Since $\boldsymbol{G_s}$ and $\boldsymbol{G_d}$ are constrained to act as auto-encoders in their own domains we incorporate classifier $C$ which trains \simplf\ and \compl\ to transfer the input sentence without auto-encoding. The losses for \simplf\ and \compl\ pipeline are mentioned below. Let $\thetaC\ $ be parameters of $C$.
% \vspace{-5pt}
\begin{align}
\small
\mathcal{L}_{div,C}(\thetaC) = -\mathbb{E}_{X_s\boldsymbol{ \sim  \mathcal{S}}}[\log{(\boldsymbol{C}(A_s(X_s)))}] - \nonumber \\
\small
\mathbb{E}_{X_d\boldsymbol{ \sim  \mathcal{D}}}[\log{(1-\boldsymbol{C}(A_d(X_d)))}]\\
\small
\mathcal{L}_{div,G_s}(\thetaE,\thetaGs) = -\mathbb{E}_{X_d\boldsymbol{ \sim  \mathcal{D}}}[\log{(\boldsymbol{C}(A_s(X_d)))}] 
\label{eqn:clfloss}
\end{align}
% \vspace{-0.25in}
\subsection{Denoising}
\label{sec:denoising}
Denoising has proven to be helpful to learn syntactic / structural transformation from the source side to the target side \cite{artetxe2018iclr}. Syntactic transformation often requires reordering the input, which the denoising procedure aims to capture. Denoising involves arbitrarily reordering the inputs and reconstructing the original (unperturbed) input from such reordered inputs. In our implementation, the source sentence is reordered by swapping bigrams in the input sentences. The following loss function are used in denoising. Let $P_{\simplfm}(X|noise(X))$ and $P_{\complm}(X|noise(X))$ denote the probabilities that a perturbed input $X$ can be reconstructed by $\simplfm$ and $\complm$ respectively. Denoising loss for models \simplf\ and \compl\ , denoted by $\mathcal{L}_{denoi}(\thetaE,\thetaGs, \thetaGd)$ is computed as follows.
\begin{align}
\footnotesize
 \mathcal{L}_{denoi} = -\mathbb{E}_{X_s\boldsymbol{ \sim  \mathcal{S}}}[\log{P_{\simplfm}(X_s|noise(X_s)})] - \nonumber \\
\small
 \mathbb{E}_{X_d\boldsymbol{ \sim  \mathcal{D}}}[\log{P_{\complm}(X_d|noise(X_d))}]
 \label{eqn:denoising}
\end{align}
%%%%%%%%%%%%Unsupervised Algorithm%%%%%%%%%%%%%%
\begin{algorithm}[t]
\caption{Unsupervised simplification algorithm using denoising, reconstruction, adversarial and diversification losses. }
\label{algo:algo1}
\textbf{Input:} simple dataset $\boldsymbol{\mathcal{{S}}}$, complex dataset $\boldsymbol{\mathcal{{D}}}$.
\begin{algorithmic}
\State \textbf{\textit{Initialization phase:}}
\Repeat
\State Update $\boldsymbol{\theta_E}$, $\boldsymbol{\theta_{G_s}}$, $\boldsymbol{\theta_{G_d}}$ using $\mathcal{L}_{denoi}$
\State Update $\boldsymbol{\theta_E}$, $\boldsymbol{\theta_{G_s}}$, $\boldsymbol{\theta_{G_d}}$ using $\mathcal{L}_{rec}$
\State Update $\boldsymbol{\theta_D}$, $\boldsymbol{\theta_C}$ using $\mathcal{L}_{adv,D}$ $\mathcal{L}_{div,C}$
\Until{specified number of steps are completed}
\State \textbf{\textit{Adversarial phase:}}
\Repeat
\State Update $\boldsymbol{\theta_E}$, $\boldsymbol{\theta_{G_s}}$, $\boldsymbol{\theta_{G_d}}$ using $\mathcal{L}_{denoi}$
\State Update $\boldsymbol{\theta_E},\boldsymbol{\theta_{G_s}},\boldsymbol{\theta_{G_d}}$ using $\mathcal{L}_{adv,G_s}$, $\mathcal{L}_{div,G_s}$, $\mathcal{L}_{rec}$
\State Update $\boldsymbol{\theta_D}$, $\boldsymbol{\theta_C}$ using $\mathcal{L}_{adv,D}$, $\mathcal{L}_{div,C}$
\Until{specified number of steps are completed}
\end{algorithmic}
\end{algorithm}
%%%%%%%%%%%%%%%%%%%%%%%%%%%%%%%%%%%%%%%%%
% %%%%%%%%%%%%Semi-supervised Algorithm%%%%%%%%%%%%%%
% \begin{algorithm}[t]
% \caption{ Semi-supervised simplification algorithm using denoising, reconstruction, adversarial and classifier losses. }
% \label{algo:algo2}
% \textbf{Input:} simple dataset $\boldsymbol{\mathcal{{S}}}$, complex dataset $\boldsymbol{\mathcal{{D}}}$, supervised corpus
% \begin{algorithmic}
% \State Define cross entropy loss $\mathcal{L}_{cross}$ using supervised corpus to train $\boldsymbol{E}$ and $\boldsymbol{G_s}$ 
% \State Train $\boldsymbol{E}$, $\boldsymbol{G_s}$, $\boldsymbol{G_d}$ using $\mathcal{L}_{denoi}$, $\mathcal{L}_{rec}$
% \State Train $\boldsymbol{E}$, $\boldsymbol{G_s}$ using $\mathcal{L}_{cross}$
% \State Train $\boldsymbol{D}$, $\boldsymbol{C}$ using $\mathcal{L}_{adv,D}$ $\mathcal{L}_{classf,C}$.

% \Repeat
% \State Train $\boldsymbol{E}$, $\boldsymbol{G_s}$, $\boldsymbol{G_d}$ using $\mathcal{L}_{denoi}$, $\mathcal{L}_{rec}$
% \State Train $\boldsymbol{E}$, $\boldsymbol{G_s}$ using $\mathcal{L}_{cross}$
% \State Train $\boldsymbol{E}$, $\boldsymbol{G_s}$ using $\mathcal{L}_{adv,G_s},\ \mathcal{L}_{classf,G_s}$
% \State Train $\boldsymbol{D}$, $\boldsymbol{C}$ using $\mathcal{L}_{adv,D}$, $\mathcal{L}_{classf,C}$.
% \Until{20,000 batches are used}
% \end{algorithmic}
% \end{algorithm}
% %%%%%%%%%%%%%%%%%%%%%%%%%%%%%%%%%%%%%%%%%
% \vspace{-0.25in}
Figure \ref{fig:workflow} depicts the overall architecture and the losses described above; the training procedure is described in Algorithm \ref{algo:algo1}. The \textit{initialization phase} involves training the \simplf, \compl~ using the reconstruction and denoising losses only. Next, training of $\boldsymbol{D}$ and \C\ happens using the respective adversarial or diversification losses. These losses are not used to update the decoders at this point. This gives the discriminator, classifier and decoders time to learn independent of each other. In the \textit{adversarial phase}, adversarial and diversification losses are introduced alongside denoising and reconstruction losses for fine-tuning the encoder and decoders. Algorithm \ref{algo:algo1} is intended to produce the following results: i) $\boldsymbol{E-G_s}$ should simplify its input (irrespective of whether it is simple or complex), and
ii) $\boldsymbol{E-G_d}$ should act as an auto-encoder in complex sentence domain. The discriminator and classifier enables preserving the appropriate aspects of semantics necessary for each of these pathways through proper modulation of the attention context vectors. 
% to extract the set of complex attention vectors $\boldsymbol{\mathcal{D}_a}$, which is required in implementing the classifier loss.

A key requirement for a model like ours is that the dataset used has to be partitioned into two sets, containing relatively simple and complex sentences. The rationale behind having two decoders is that while $\boldsymbol{G_s}$ will try to introduce simplified constructs (may be at the expense of loss of semantics), $\boldsymbol{G_d}$ will help preserve the semantics. The idea behind using the discriminator and classifier is to retain signals related to language simplicity from which $\boldsymbol{G_s}$ will construct simplified sentences. Finally, denoising will help tackle nuances related to syntactic transfer from complex to simple direction. We remind the readers that, TS, unlike machine translation, needs complex syntactic operations such as sentence splitting, rephrasing and paraphrasing, which can not be tackled by the losses and denoising alone. Employing additional explicit mechanisms to handle these in the pipeline is out of the scope of this paper since we seek a \textit{prima-facie} judgement of our architecture based on how much simplification knowledge can be gained just from the data.
\subsection{Training with Minimal Supervision}
\label{sec:semisup}
Our system, by design, is highly data-driven, and like any other sequence-to-sequence learning based system, can also leverage labeled data. We propose a semi-supervised variant of our system that could gain additional knowledge of simplification through the help of a small amount of labeled data (in the order of a few thousands). The system undergoes training following steps similar to Algorithm \ref{algo:algo1}, except that it adds another step of optimizing the \textit{cross entropy loss} for both the \simplf\ and \compl\ pathways by using the reference texts available in the labeled dataset. This step is carried out in the adversarial phase along with other steps (See Algorithm \ref{algo:algo2}).
%in Appendix \ref{sec:app}).

The cross-entropy loss is imposed on both \simplf\ and \compl\ paths using parallel dataset (details mentioned in Section \ref{subsec:dataset}) denoted by $\Delta = (\boldsymbol{\mathcal{S}_p}, \boldsymbol{\mathcal{D}_p})$.
For a given parallel simplification sentence pair $(X_s,X_d)$,
let $P_{\boldsymbol{E-G_s}}({X_s}|{X_d})$ and $P_{\boldsymbol{E-G_d}}(X_d|X_s)$ denote the probabilities that $X_s$ is produced from $X_d$  by the \simplf\ and the reverse is produced by the \compl\ respectively. Cross-Entropy loss for \simplf\ and \compl\ denoted by $\mathcal{L}_{cross}(\thetaE,\thetaGs,\thetaGd)$ is computed as follows:
% \vspace{-5pt}
\begin{align}
\small
 \mathcal{L}_{cross} = -\mathbb{E}_{(X_s,X_d)\boldsymbol{ \sim  \Delta}}[\log{P_{\simplfm}(X_s|X_d)}] - \nonumber \\
\small
 \mathbb{E}_{(X_s,X_d)\boldsymbol{ \sim  \Delta}}[\log{P_{\complm}(X_d|X_s)}]
\label{eqn:reconst}
\end{align}

\begin{algorithm}[h]
\caption{Semi-supervised simplification algorithm using denoising, reconstruction, adversarial and diversification losses followed by cross-entropy loss using parallel data. }
\label{algo:algo2}
\textbf{Input:} simple dataset $\boldsymbol{\mathcal{{S}}}$, complex dataset $\boldsymbol{\mathcal{{D}}}$, parallel dataset $\Delta = (\boldsymbol{\mathcal{S}_p}, \boldsymbol{\mathcal{D}_p})$
\begin{algorithmic}
\State \textbf{\textit{Initialization phase:}}
\Repeat
\State Update $\boldsymbol{\theta_E}$, $\boldsymbol{\theta_{G_s}}$, $\boldsymbol{\theta_{G_d}}$ using $\mathcal{L}_{denoi}$
\State Update $\boldsymbol{\theta_E}$, $\boldsymbol{\theta_{G_s}}$, $\boldsymbol{\theta_{G_d}}$ using $\mathcal{L}_{rec}$
\State Update $\boldsymbol{\theta_D}$, $\boldsymbol{\theta_C}$ using $\mathcal{L}_{adv,D}$ $\mathcal{L}_{div,C}$
\Until{specified number of steps are completed}
\State \textbf{\textit{Adversarial phase:}}
\Repeat
\State Update $\boldsymbol{\theta_E}$, $\boldsymbol{\theta_{G_s}}$, $\boldsymbol{\theta_{G_d}}$ using $\mathcal{L}_{denoi}$
\State Update $\boldsymbol{\theta_E},\boldsymbol{\theta_{G_s}},\boldsymbol{\theta_{G_d}}$ using $\mathcal{L}_{adv,G_s}$, $\mathcal{L}_{div,G_s}$, $\mathcal{L}_{rec}$
\State Update $\boldsymbol{\theta_D}$, $\boldsymbol{\theta_C}$ using $\mathcal{L}_{adv,D}$, $\mathcal{L}_{div,C}$
\State Update $\boldsymbol{\theta_E}$, $\boldsymbol{\theta_{G_s}}$ using $\mathcal{L}_{cross}$
\State Update $\boldsymbol{\theta_E}$, $\boldsymbol{\theta_{G_d}}$ using $\mathcal{L}_{cross}$
\Until{specified number of steps are completed}
\end{algorithmic}
\end{algorithm}

\section{Experiment Setup}
\label{sec:exp}
In this section we describe the dataset, architectural choices, and model hyperparameters. The implementation of the experimental setup is publicly available\footnote{https://github.com/subramanyamdvss/UnsupNTS}.
\begin{table}[t]
\centering
\footnotesize
\begin{tabular}{lllll}
    \toprule
    \textbf{Category} & \textbf{\#Sents} & \textbf{Avg.} & \textbf{Avg.} & \textbf{FE-} \\
     & & \textbf{Words} & \textbf{FE} & \textbf{Range} \\
    \midrule
    Simple  & 720k & 18.23 & 76.67 & 74.9-79.16\\
    Complex & 720k & 35.03 & 7.26 & 5.66-9.93\\
    \bottomrule
\end{tabular}
\caption{Statistics showing number of sentences, average words per sentence, and average FE score, FE score limits for complex and simple datasets used for training.}
\label{tab:datastat}
\end{table}
\subsection{Dataset}
\label{subsec:dataset}
For training our system, we created an unlabeled dataset of simple and complex sentences by partitioning the standard \textit{en-wikipedia} dump. Since partitioning requires a metric for measuring text simpleness we categorize sentences based on their readability scores. For this we use the Flesch Readability Ease (henceforth abbreviated as FE) \cite{flesch1948new}. Sentences with lower FE values (up to 10) are categorized as complex and sentences with FE values greater than 70 are categorized as simple\footnote{FE has its shortcomings to fully judge simpleness, but we nevertheless employ it in the absence of stronger metrics}. The FE bounds are decided through trial and error through manual inspection of the categorized sentences. Table \ref{tab:datastat} shows dataset statistics. Even though the dataset was created with some level of human mediation, the manual effort is insignificant compared to that needed to create a parallel  corpus. 

To train the system with minimal supervision (Section \ref{sec:semisup}), we extract $10,000$ pairs of sentences from various datasets such as Wikipedia-SimpleWikipedia dataset introduced in \newcite{hwang2015aligning} and the Split-Rephrase dataset by \newcite{narayan2017split}\footnote{https://github.com/shashiongithub/Split-and-Rephrase}. The Wikipedia-SimpleWikipedia was filtered following \newcite{nisioi2017exploring} and $4000$ examples were randomly picked from the filtered set. From the Split-Rephrase dataset, examples containing one compound/complex sentence at the source side and two simple sentences at the target side were selected and $6000$ examples were randomly picked from the selected set. The Split-Rephrase dataset is used to promote sentence splitting in the proposed system.

To select and evaluate our models, we use the test and development sets\footnote{We acknowledge that other recent datasets such as Newsela could have been used for development and evaluation. We could not get access to the dataset unfortunately.} released by \cite{xu2016optimizing}. The test set ($359$ sentences) and development set ($2000$ sentences) have 8 simplified reference sentences for each source sentence. 

\subsection{Hyperparameter Settings}
\label{subsec:param}
For all the variants, we use a hidden state of size 600 and word-embedding size of 300. Classifier $C$ and discriminator $D$ use convolutional layers with filters sizes from $1$ to $5$. 128 filters of each size are used in the CNN-layers. Other training related hyper parameters include learning rate of 0.00012 for $\boldsymbol{\theta_E,\theta_{G_s},\theta_{G_d}}$ , 0.0005 for $\boldsymbol{\theta_D,\theta_C}$ and batch size of 36. For learning the word-embedding using \textit{Dict2Vec} training, the window size is set to 5. Our experiments used at most 13 GB of GPU memory. The Initialization phase and Adversarial phase took 6000 and 8000 steps in batches respectively for both \textsc{UNTS} and \textsc{UNTS}+10K systems.

\subsection{Evaluation Metrics}
For automatic evaluation of our system on the test data, we used four metrics, (a) SARI (b)  BLEU (c) FE Difference (d) Word Difference, which are briefly explained below.

SARI \cite{xu2016optimizing} is an automatic evaluation metric designed to measure the simpleness of the generated sentences. SARI requires access to source, predictions and references for evaluation. Computing SARI involves penalizing the n-gram additions to source which are inconsistent with the references. Similarly, deletions and keep operations are penalized. The overall score is a balanced sum of all the penalties. BLEU \cite{papineni2002bleu}, a popular metric to evaluate generations and translations is used to measure the correctness of the generations by measuring overlaps between the generated sentences and (multiple) references.

We also compute the average FE score difference between predictions and source in our evaluations. FE-difference measures whether the changes made by the model increase the readability ease of the generated sentence. \textit{Word Difference} is the average difference between number of words in the source sentence and generation. It is a simple and approximate metric proposed to detect if sentence shortening is occurring or not. Generations with lesser number of changes can still have high SARI and BLEU. Models with such generations can be ruled out by imposing a threshold on the word-diff metric.

Models with high word-diff, SARI and BLEU are picked during model-selection (with validation data). Model selection also involved manually examining the quality and relevance of generations.

We carry out a qualitative analysis of our system through human evaluation. For this the first 50 test samples were selected from the test data. Output of the seven systems reported in Table \ref{tab:results} along with the sources are presented to two native English speakers who would provide two ratings for each output: (a) \textit{Simpleness}, a binary score [0-1] indicating whether the output is a simplified version of the input or not, (b) \textit{Grammaticality} of the output in the range of [1-5], in the increasing order of fluency (c) \textit{Relatedness} score in the range of [1-5] showing if the overall semantics of the input is preserved in the output or not.

\subsection{Model Variants}
Using our design, we propose two different variants for evaluation: (i) Unsupervised Neural TS (\textsc{UNTS}) with SARI as the criteria for model selection, (ii) \textsc{UNTS} with minimal supervision using $10000$ labelled examples (UNTS+10K). Models selected using other selection criteria such as BLEU resulted in similar and/or reduced performance (details skipped for brevity).

We carried out the following basic post-processing steps on the generated outputs. The OOV(out of vocabulary) words in the generations are replaced by the source words with high attention weights. Words repeated consecutively  in the generated sentences are merged.

\subsection{Systems for Comparison}
\label{exp:exp}
In the absence of any other direct baseline for end-to-end TS, we consider the following unsupervised baselines. We consider the unsupervised NMT framework proposed by \cite{artetxe2018iclr} as a baseline. It uses techniques such as \textit{backtranslation} and \textit{denoising} techniques to synthesize more training examples. To use this framework, we treated the set of simple and complex sentences as two different languages. Same model configuration as reported by  \newcite{artetxe2018iclr} is used. We use the term \textsc{UNMT} for this system. 

Similar to the \textsc{UNMT} system, we also consider unsupervised statistical machine translation (termed as \textsc{USMT}) proposed by \newcite{artetxe2018unsupervised}, with default parameter setting. Another system, based on the cross alignment technique proposed by \newcite{shen2017style} is also used for comparison. The system is originally proposed for the task of sentiment translation. We term this system as \textsc{ST}.

We also compare our approach with existing supervised and unsupervised lexical simplifications like \textsc{LightLS} \cite{glavavs2015simplifying}, Neural Text Simplification or \textsc{NTS} \cite{nisioi2017exploring}, Syntax based Machine Translation or \textsc{SbMT} \cite{xu2016optimizing}, and Phrase-based SMT simplification or \textsc{PbSMT} \cite{wubben2012sentence}. All the systems are trained using the Wikipedia-SimpleWikipedia dataset \cite{hwang2015aligning}. The test set is same for all of these and our models.
\section{Results}
\label{sec:result}
\begin{table}
\footnotesize
\begin{tabular}{lllll}
\toprule
\textbf{System} & \textbf{FE-diff} & \textbf{SARI} & \textbf{BLEU} & \textbf{Word-diff} \\
\midrule
\textsc{UNTS}+10K & 10.45 & 35.29 & 76.13 & 2.38 \\ 
\textsc{UNTS} & 11.15 & 33.8 & 74.24 & \textbf{3.55} \\
\midrule
\textsc{UNMT} & 6.60 & 33.72 & 70.84 & 0.74 \\
\textsc{USMT} & 13.84 & 32.11 & 87.36 & -0.01\\
\textsc{ST} & 54.38 & 14.97 & 0.73 & 5.61\\
\midrule
% NTS-BLEU & 2.84 & 30.77 & 84.7 & 2.47 \\
\textsc{NTS} & 5.37 & 36.1 & 79.38 & 2.73 \\
\textsc{SbMT} & \textbf{17.68} & \textbf{38.59} & 73.62 & -0.84 \\
\textsc{PbSMT} & 9.14 & 34.07 & 67.79 & 2.26 \\
\midrule
\textsc{LightLS} & 3.01 & 34.96 & \textbf{83.54} & -0.02 \\ \bottomrule
\end{tabular}
\caption{Comparison of evaluation metrics for proposed systems (\textsc{UNTS}), unsupervised baseline (\textsc{UNMT},\textsc{USMT}, and \textsc{ST}) and existing supervised and the unsupervised lexical simplification system \textsc{LightLS}.}
\label{tab:results} 
\end{table}
\begin{table}[t]
\footnotesize
\begin{tabular}{llll}
\toprule
\textbf{System} & \textbf{Simpleness} & \textbf{Fluency} & \textbf{Relatedness}\\
\midrule
\textsc{UNTS}+10K & \textbf{57}\% & 4.13 & 3.93 \\ 
\textsc{UNTS} & 47\% & 3.86 & 3.73 \\
\midrule
\textsc{UNMT} & 40\% & 3.8 & \textbf{4.06} \\
\midrule
\textsc{NTS} & 49\% & 4.13 & 3.26 \\
\textsc{SbMT} & 53\% & \textbf{4.26} & \textbf{4.06} \\
\textsc{PbSMT} & 53\% & 3.8 & 3.93 \\
\midrule
\textsc{LightLS} & 6\% & 4.2 & 3.33 \\ \bottomrule
\end{tabular}
\caption{Average human evaluation scores for simpleness and grammatical correctness (fluency) and semantic relatedness between the output and input.}
\label{tab:Human} 
\end{table}
\begin{table*}[t]
\scriptsize
\begin{tabular}{p{3cm} p{12cm}}
\toprule
\textbf{System} & \textbf{Output} \\
\midrule
\textbf{Input} & Nevertheless , Tagore emulated numerous styles , including craftwork from northern New Ireland , Haida carvings from the west coast of Canada ( British Columbia ) , and woodcuts by Max Pechstein .\\
\textbf{Reference} & Nevertheless , Tagore copied many styles , such as crafts from northern New Ireland , Haida carvings from the west coast of Canada and wood carvings by Max Pechstein .\\
\midrule 
\textsc{UNTS}+10K & Nevertheless , Tagore replaced many styles , including craftwork from northern New Ireland , Haida carved from the west coast of Canada ( British Columbia ) . \\ \textsc{UNTS} & However , Tagore notably numerous styles , including craftwork from northern New Ireland , Haida carved from the west coast of Canada ( British ) . \\ \midrule
\textsc{UNMT} & However , Tagore featured numerous styles including craftwork from northern New Ireland , Haida from the west coast of Canada ( British Columbia ) max by Max Pechstein .\\
\textsc{USMT} & Nevertheless , Mgr emulated numerous styles , including craftwork from northern New Ireland , Haida carvings from the west coast of Canada (British Columbia) , and etchings by Max Pechstein .\\
\midrule
\textsc{NTS} & However , Tagore wrote many styles , including craftwork from northern New Ireland , Haida carvings from the west coast of Canada ( British Columbia ) . \\
\textsc{SbMT} & However , Tagore emulated many styles , such as craftwork in north New Ireland , Haida prints from the west coast of Canada ( British Columbia ) , and woodcuts by Max Pechstein . \\
\textsc{PbSMT} & Nevertheless , he copied many styles , from new craftwork , Haida carvings from the west coast of Canada in British Columbia and woodcuts by Max Pechstein . \\
\midrule
\textsc{LightLS} & However , Tagore imitated numerous styles , including craftwork from northern New Ireland , Haida sculptures from the west coast of Canada ( British Columbia ) , and engravings by Max Pechstein . \\
\bottomrule
\end{tabular}
\caption{Example predictions from different systems.}
\label{tab:gens}
\end{table*}

\begin{table*}[t]
\scriptsize

\begin{tabular}{p{2.28cm}p{6.1cm}p{6.2cm}}
\toprule
\textbf{Type of Simplification} & \textbf{Source} & \textbf{Prediction} \\ 
\midrule
% Splitting & Gumbasia was created in a style Vorkapich taught called Kinesthetic Film Principles . & Gumbasia was created in Vorkapich . \textbf{Vorkapich} is a style of Kinesthetic Principles . \\ \hline
Splitting & Calvin Baker is an American novelist . & Calvin Baker is an American . American Baker is a birthplace .\\ \midrule
Sentence Shortening & During an interview , Edward Gorey mentioned that Bawden was one of his favorite artists , \textbf{lamenting the fact that not many people remembered or knew about this fine artist .} & During an interview , Edward Gorey mentioned that Bawden was one of his favorite artists . \\ \midrule
Lexical Replacement & In \textbf{architectural} decoration Small pieces of colored and \textbf{iridescent} shell have been used to create \textbf{mosaics} and inlays , which have been used to decorate walls , furniture and boxes . & In \textbf{impressive} decoration Small pieces of colored and \textbf{reddish} shell have been used to create \textbf{statues} and inlays , which have been used to decorate walls , furniture and boxes . \\ \bottomrule
\end{tabular}
\caption{Examples showing different types of simplifications performed by the best model \textsc{UNTS}+10K.}
\label{tab:typesimplf}
\end{table*}
Table \ref{tab:results} shows evaluation results of our proposed approaches along with existing supervised and unsupervised alternatives. We observe that unsupervised baselines such as \textsc{UNMT} and \textsc{USMT} often, after attaining convergence, recreates sentences similar to the inputs. This explains why they achieve higher BLEU and reduced word-difference scores. The \textsc{ST} system did not converge for our dataset after significant number of epochs which affected the performance metrics. The system often produces short sentences which are simple but do not retain important phrases. 

Other supervised systems such as \textsc{SbMT} and \textsc{NTS} achieve better content reduction as shown through SARI, BLEU and FE-diff scores; this is expected. However, it is still a good sign that the scores for the unsupervised system \textsc{UNTS} are not far from the supervised skylines. The higher word-diff scores for the unsupervised system also indicate that it is able to perform content reduction (a form of syntactic simplification), which is crucial to TS. This is unlike the existing unsupervised \textsc{LightLS} system which often replaces nouns with related non-synonymous nouns; sometimes increasing the complexity and affecting the meaning. Finally, it is worth noting that aiding the system with a very small amount of labeled data can also benefit our unsupervised pipeline, as suggested by the scores for the  \textsc{UNTS}+10K system. 

In Table \ref{tab:Human}, the first column represents what percentage of output form is a simplified version of the input. The second and third columns present the average fluency (grammaticality) scores given by human evaluators and semantic relatedness with input scored through automatic means. Almost all systems are able to produce sentences that are somewhat grammatically correct and retain phrases from input. Supervised systems like \textsc{PbSMT}, as expected, simplify the sentences to the maximum extent. However, our unsupervised variants have scores competitive to the supervised skylines, which is a positive sign.

Table \ref{tab:gens} shows an anecdotal example, containing outputs from the seven systems. As can be seen, the quality of output from our unsupervised variants, is far from that of the reference output. However, the attempts towards performing lexical simplification (by replacing the word ``Neverthless'' with ``However'') and simplification of multi-word phrases (``Tagore emulated numerous styles''
getting translated to ``Tagore replaced many styles'') are quite visible and encouraging. Table \ref{tab:typesimplf} presents a few examples demonstrating the capabilities of our system in performing simplifications at lexical and syntactic level. We do observe that such operations are carried out only for a few instances in our test data. Also, our analysis in Appendix \ref{appendix:datavar}  indicate that the system can improve over time with addition of more data. Results for ablations on adversarial and diversification loss are also included in Appendix \ref{appendix:ablation}. 
\section{Conclusion}
\label{sec:concl}
In this paper, we made a novel attempt towards unsupervised text simplification. We gathered unlabeled corpora containing simple and complex sentences and used them to train our system that is based on a shared encoder and two decoders. A novel training scheme is proposed which allows the model to perform content reduction and lexical simplification simultaneously through our proposed losses and denoising. Experiments were conducted for multiple variants of our system as well as known unsupervised baselines and supervised systems. Qualitative and quantitative analysis of the outputs for a publicly available test data demonstrate that our models, though unsupervised, can perform better than or competitive to these baselines. In future, we would like to improve the system further by incorporating better architectural designs and training schemes to tackle complex simplification operations. %Applying our models for multi- and cross-lingual TS is is also on our agenda.
\section{Acknowledgements}
We thank researchers at IBM IRL, IIT Kharagpur, Vishal Gupta and Dr. Sudeshna Sarkar for helpful discussions in this project.
%%%%%%%%%%%Sectiowise input %%%%%%%%%%%
% \begin{spacing}{0.1}

\bibliography{acl2019}

\begin{thebibliography}{46}
\expandafter\ifx\csname natexlab\endcsname\relax\def\natexlab#1{#1}\fi

\bibitem[{Artetxe et~al.(2018{\natexlab{a}})Artetxe, Labaka, and
  Agirre}]{artetxe2018unsupervised}
Mikel Artetxe, Gorka Labaka, and Eneko Agirre. 2018{\natexlab{a}}.
\newblock Unsupervised statistical machine translation.
\newblock In \emph{Proceedings of the 2018 Conference on Empirical Methods in
  Natural Language Processing}, pages 3632--3642.

\bibitem[{Artetxe et~al.(2018{\natexlab{b}})Artetxe, Labaka, Agirre, and
  Cho}]{artetxe2018iclr}
Mikel Artetxe, Gorka Labaka, Eneko Agirre, and Kyunghyun Cho.
  2018{\natexlab{b}}.
\newblock Unsupervised neural machine translation.
\newblock In \emph{Proceedings of the Sixth International Conference on
  Learning Representations}.

\bibitem[{Bahdanau et~al.(2014)Bahdanau, Cho, and Bengio}]{bahdanau2014neural}
Dzmitry Bahdanau, Kyunghyun Cho, and Yoshua Bengio. 2014.
\newblock Neural machine translation by jointly learning to align and
  translate.
\newblock \emph{arXiv:1409.0473}.

\bibitem[{Biran et~al.(2011)Biran, Brody, and Elhadad}]{biran2011putting}
Or~Biran, Samuel Brody, and No{\'e}mie Elhadad. 2011.
\newblock Putting it simply: a context-aware approach to lexical
  simplification.
\newblock In \emph{ACL}, pages 496--501. Association for Computational
  Linguistics.

\bibitem[{Brouwers et~al.(2014)Brouwers, Bernhard, Ligozat, and
  Fran{\c{c}}ois}]{brouwers2014syntactic}
Laetitia Brouwers, Delphine Bernhard, Anne-Laure Ligozat, and Thomas
  Fran{\c{c}}ois. 2014.
\newblock Syntactic sentence simplification for french.
\newblock In \emph{Proceedings of the 3rd Workshop on Predicting and Improving
  Text Readability for Target Reader Populations (PITR)@ EACL 2014}, pages
  47--56.

\bibitem[{Candido~Jr et~al.(2009)Candido~Jr, Maziero, Gasperin, Pardo, Specia,
  and Aluisio}]{candido2009supporting}
Arnaldo Candido~Jr, Erick Maziero, Caroline Gasperin, Thiago~AS Pardo, Lucia
  Specia, and Sandra~M Aluisio. 2009.
\newblock Supporting the adaptation of texts for poor literacy readers: a text
  simplification editor for brazilian portuguese.
\newblock In \emph{Innovative Use of NLP for Building Educational
  Applications}, pages 34--42. Association for Computational Linguistics.

\bibitem[{Canning and Tait(1999)}]{canning1999syntactic}
Yvonne Canning and John Tait. 1999.
\newblock Syntactic simplification of newspaper text for aphasic readers.
\newblock In \emph{Customised Information Delivery}, pages 6--11.

\bibitem[{Chandrasekar and Srinivas(1997)}]{chandrasekar1997automatic}
Raman Chandrasekar and Bangalore Srinivas. 1997.
\newblock Automatic induction of rules for text simplification1.
\newblock \emph{Knowledge-Based Systems}, 10(3):183--190.

\bibitem[{Cho et~al.(2014{\natexlab{a}})Cho, van Merrienboer, Bahdanau, and
  Bengio}]{cho2014properties}
Kyunghyun Cho, Bart van Merrienboer, Dzmitry Bahdanau, and Yoshua Bengio.
  2014{\natexlab{a}}.
\newblock On the properties of neural machine translation: Encoder--decoder
  approaches.
\newblock In \emph{Proceedings of SSST-8, Eighth Workshop on Syntax, Semantics
  and Structure in Statistical Translation}, pages 103--111.

\bibitem[{Cho et~al.(2014{\natexlab{b}})Cho, van Merrienboer, Gulcehre,
  Bahdanau, Bougares, Schwenk, and Bengio}]{cho2014learning}
Kyunghyun Cho, Bart van Merrienboer, Caglar Gulcehre, Dzmitry Bahdanau, Fethi
  Bougares, Holger Schwenk, and Yoshua Bengio. 2014{\natexlab{b}}.
\newblock Learning phrase representations using rnn encoder--decoder for
  statistical machine translation.
\newblock In \emph{Proceedings of the 2014 Conference on Empirical Methods in
  Natural Language Processing (EMNLP)}, pages 1724--1734.

\bibitem[{Clarke and Lapata(2006)}]{clarke2006models}
James Clarke and Mirella Lapata. 2006.
\newblock Models for sentence compression: A comparison across domains,
  training requirements and evaluation measures.
\newblock In \emph{COLING}, pages 377--384. Association for Computational
  Linguistics.

\bibitem[{Coster and Kauchak(2011)}]{coster2011simple}
William Coster and David Kauchak. 2011.
\newblock Simple english wikipedia: a new text simplification task.
\newblock In \emph{ACL}, pages 665--669. Association for Computational
  Linguistics.

\bibitem[{Devlin(1998)}]{devlin1998use}
Siobhan Devlin. 1998.
\newblock The use of a psycholinguistic database in the simplification of text
  for aphasic readers.
\newblock \emph{Linguistic databases}.

\bibitem[{Filippova et~al.(2015)Filippova, Alfonseca, Colmenares, Kaiser, and
  Vinyals}]{filippova2015sentence}
Katja Filippova, Enrique Alfonseca, Carlos~A Colmenares, Lukasz Kaiser, and
  Oriol Vinyals. 2015.
\newblock Sentence compression by deletion with lstms.
\newblock In \emph{Proceedings of the 2015 Conference on Empirical Methods in
  Natural Language Processing}, pages 360--368.

\bibitem[{Filippova and Strube(2008)}]{filippova2008dependency}
Katja Filippova and Michael Strube. 2008.
\newblock Dependency tree based sentence compression.
\newblock In \emph{INLG}, pages 25--32. Association for Computational
  Linguistics.

\bibitem[{Flesch(1948)}]{flesch1948new}
Rudolph Flesch. 1948.
\newblock A new readability yardstick.
\newblock \emph{Journal of applied psychology}, 32(3):221.

\bibitem[{Glava{\v{s}} and {\v{S}}tajner(2015)}]{glavavs2015simplifying}
Goran Glava{\v{s}} and Sanja {\v{S}}tajner. 2015.
\newblock Simplifying lexical simplification: do we need simplified corpora?
\newblock In \emph{ACL}, volume~2, pages 63--68.

\bibitem[{Goodfellow et~al.(2014)Goodfellow, Pouget-Abadie, Mirza, Xu,
  Warde-Farley, Ozair, Courville, and Bengio}]{goodfellow2014generative}
Ian Goodfellow, Jean Pouget-Abadie, Mehdi Mirza, Bing Xu, David Warde-Farley,
  Sherjil Ozair, Aaron Courville, and Yoshua Bengio. 2014.
\newblock Generative adversarial nets.
\newblock In \emph{Advances in neural information processing systems}, pages
  2672--2680.

\bibitem[{Hwang et~al.(2015)Hwang, Hajishirzi, Ostendorf, and
  Wu}]{hwang2015aligning}
William Hwang, Hannaneh Hajishirzi, Mari Ostendorf, and Wei Wu. 2015.
\newblock Aligning sentences from standard wikipedia to simple wikipedia.
\newblock In \emph{NAACL-HLT}, pages 211--217.

\bibitem[{Kim(2014)}]{kim2014convolutional}
Yoon Kim. 2014.
\newblock Convolutional neural networks for sentence classification.
\newblock \emph{arXiv:1408.5882}.

\bibitem[{Knight and Marcu(2002)}]{knight2002summarization}
Kevin Knight and Daniel Marcu. 2002.
\newblock Summarization beyond sentence extraction: A probabilistic approach to
  sentence compression.
\newblock \emph{Artificial Intelligence}, 139(1):91--107.

\bibitem[{L'Allier(1980)}]{lallier80eval}
J.~L'Allier. 1980.
\newblock An evaluation study of a computer-based lesson that adjusts read- ing
  level by monitoring on task reader characteristics.
\newblock \emph{Ph.D. Thesis}.

\bibitem[{Linderholm et~al.(2000)Linderholm, Everson, Van Den~Broek,
  Mischinski, Crittenden, and Samuels}]{linderholm2000effects}
Tracy Linderholm, Michelle~Gaddy Everson, Paul Van Den~Broek, Maureen
  Mischinski, Alex Crittenden, and Jay Samuels. 2000.
\newblock Effects of causal text revisions on more-and less-skilled readers'
  comprehension of easy and difficult texts.
\newblock \emph{Cognition and Instruction}, 18(4):525--556.

\bibitem[{McNamara et~al.(1996)McNamara, Kintsch, Songer, and
  Kintsch}]{mcnamara1996good}
Danielle~S McNamara, Eileen Kintsch, Nancy~Butler Songer, and Walter Kintsch.
  1996.
\newblock Are good texts always better? interactions of text coherence,
  background knowledge, and levels of understanding in learning from text.
\newblock \emph{Cognition and instruction}, 14(1):1--43.

\bibitem[{Narayan and Gardent(2014)}]{narayan2014hybrid}
Shashi Narayan and Claire Gardent. 2014.
\newblock Hybrid simplification using deep semantics and machine translation.
\newblock In \emph{ACL}, volume~1, pages 435--445.

\bibitem[{Narayan et~al.(2017)Narayan, Gardent, Cohen, and
  Shimorina}]{narayan2017split}
Shashi Narayan, Claire Gardent, Shay Cohen, and Anastasia Shimorina. 2017.
\newblock Split and rephrase.
\newblock In \emph{EMNLP 2017: Conference on Empirical Methods in Natural
  Language Processing}, pages 617--627.

\bibitem[{Nisioi et~al.(2017)Nisioi, {\v{S}}tajner, Ponzetto, and
  Dinu}]{nisioi2017exploring}
Sergiu Nisioi, Sanja {\v{S}}tajner, Simone~Paolo Ponzetto, and Liviu~P Dinu.
  2017.
\newblock Exploring neural text simplification models.
\newblock In \emph{ACL}, volume~2, pages 85--91.

\bibitem[{Paetzold and Specia(2016)}]{paetzold2016unsupervised}
Gustavo~H Paetzold and Lucia Specia. 2016.
\newblock Unsupervised lexical simplification for non-native speakers.
\newblock In \emph{AAAI}, pages 3761--3767.

\bibitem[{Papineni et~al.(2002)Papineni, Roukos, Ward, and
  Zhu}]{papineni2002bleu}
Kishore Papineni, Salim Roukos, Todd Ward, and Wei-Jing Zhu. 2002.
\newblock Bleu: a method for automatic evaluation of machine translation.
\newblock In \emph{ACL}, pages 311--318. Association for Computational
  Linguistics.

\bibitem[{Petersen and Ostendorf(2007)}]{petersen2007text}
Sarah~E Petersen and Mari Ostendorf. 2007.
\newblock Text simplification for language learners: a corpus analysis.
\newblock In \emph{Workshop on Speech and Language Technology in Education}.

\bibitem[{Qi et~al.(2018)Qi, Sachan, Felix, Padmanabhan, and
  Neubig}]{qi2018and}
Ye~Qi, Devendra Sachan, Matthieu Felix, Sarguna Padmanabhan, and Graham Neubig.
  2018.
\newblock When and why are pre-trained word embeddings useful for neural
  machine translation?
\newblock In \emph{Proceedings of the 2018 Conference of the North American
  Chapter of the Association for Computational Linguistics: Human Language
  Technologies, Volume 2 (Short Papers)}, pages 529--535.

\bibitem[{Rush et~al.(2015)Rush, Chopra, and Weston}]{rush2015neural}
Alexander~M Rush, Sumit Chopra, and Jason Weston. 2015.
\newblock A neural attention model for abstractive sentence summarization.
\newblock In \emph{Proceedings of the 2015 Conference on Empirical Methods in
  Natural Language Processing}, pages 379--389.

\bibitem[{Shen et~al.(2017)Shen, Lei, Barzilay, and Jaakkola}]{shen2017style}
Tianxiao Shen, Tao Lei, Regina Barzilay, and Tommi Jaakkola. 2017.
\newblock Style transfer from non-parallel text by cross-alignment.
\newblock In \emph{Advances in neural information processing systems}, pages
  6830--6841.

\bibitem[{Siddharthan(2006)}]{siddharthan2006syntactic}
Advaith Siddharthan. 2006.
\newblock Syntactic simplification and text cohesion.
\newblock \emph{Research on Language and Computation}, 4(1):77--109.

\bibitem[{Siddharthan(2014)}]{siddharthan2014survey}
Advaith Siddharthan. 2014.
\newblock A survey of research on text simplification.
\newblock \emph{ITL-International Journal of Applied Linguistics},
  165(2):259--298.

\bibitem[{Specia(2010)}]{specia2010translating}
Lucia Specia. 2010.
\newblock Translating from complex to simplified sentences.
\newblock In \emph{Computational Processing of the Portuguese Language}, pages
  30--39. Springer.

\bibitem[{{\v{S}}tajner et~al.(2015){\v{S}}tajner, Bechara, and
  Saggion}]{vstajner2015deeper}
Sanja {\v{S}}tajner, Hannah Bechara, and Horacio Saggion. 2015.
\newblock A deeper exploration of the standard pb-smt approach to text
  simplification and its evaluation.
\newblock In \emph{ACL-IJCNLP}, volume~2, pages 823--828.

\bibitem[{{\v{S}}tajner and Nisioi(2018)}]{stajner2018}
Sanja {\v{S}}tajner and Sergiu Nisioi. 2018.
\newblock {A Detailed Evaluation of Neural Sequence-to-Sequence Models for
  In-domain and Cross-domain Text Simplification}.
\newblock In \emph{Proceedings of the Eleventh International Conference on
  Language Resources and Evaluation (LREC 2018)}, Miyazaki, Japan. European
  Language Resources Association (ELRA).

\bibitem[{Tissier et~al.(2017)Tissier, Gravier, and
  Habrard}]{tissier2017dict2vec}
Julien Tissier, Christophe Gravier, and Amaury Habrard. 2017.
\newblock Dict2vec : Learning word embeddings using lexical dictionaries.
\newblock In \emph{EMNLP}, pages 254--263.

\bibitem[{Wang et~al.(2016)Wang, Chen, Rochford, and Qiang}]{wang2016text}
Tong Wang, Ping Chen, John Rochford, and Jipeng Qiang. 2016.
\newblock Text simplification using neural machine translation.
\newblock In \emph{AAAI}.

\bibitem[{Wubben et~al.(2012)Wubben, Van Den~Bosch, and
  Krahmer}]{wubben2012sentence}
Sander Wubben, Antal Van Den~Bosch, and Emiel Krahmer. 2012.
\newblock Sentence simplification by monolingual machine translation.
\newblock In \emph{Proceedings of the 50th Annual Meeting of the Association
  for Computational Linguistics: Long Papers-Volume 1}, pages 1015--1024.
  Association for Computational Linguistics.

\bibitem[{Xu et~al.(2018)Xu, Xu, Zeng, Zhang, Ren, Wang, and
  Li}]{xu2018unpaired}
Jingjing Xu, SUN Xu, Qi~Zeng, Xiaodong Zhang, Xuancheng Ren, Houfeng Wang, and
  Wenjie Li. 2018.
\newblock Unpaired sentiment-to-sentiment translation: A cycled reinforcement
  learning approach.
\newblock In \emph{Proceedings of the 56th Annual Meeting of the Association
  for Computational Linguistics (Volume 1: Long Papers)}, pages 979--988.

\bibitem[{Xu et~al.(2015)Xu, Callison-Burch, and Napoles}]{xu2015problems}
Wei Xu, Chris Callison-Burch, and Courtney Napoles. 2015.
\newblock Problems in current text simplification research: New data can help.
\newblock \emph{Transactions of the Association of Computational Linguistics},
  3(1):283--297.

\bibitem[{Xu et~al.(2016)Xu, Napoles, Pavlick, Chen, and
  Callison-Burch}]{xu2016optimizing}
Wei Xu, Courtney Napoles, Ellie Pavlick, Quanze Chen, and Chris Callison-Burch.
  2016.
\newblock Optimizing statistical machine translation for text simplification.
\newblock \emph{TACL}, 4:401--415.

\bibitem[{Yatskar et~al.(2010)Yatskar, Pang, Danescu-Niculescu-Mizil, and
  Lee}]{yatskar2010sake}
Mark Yatskar, Bo~Pang, Cristian Danescu-Niculescu-Mizil, and Lillian Lee. 2010.
\newblock For the sake of simplicity: Unsupervised extraction of lexical
  simplifications from wikipedia.
\newblock In \emph{NAACL-HLT}, pages 365--368. Association for Computational
  Linguistics.

\bibitem[{Zhang et~al.(2018)Zhang, Ren, Liu, Wang, Chen, Li, Zhou, and
  Chen}]{zhang2018style}
Zhirui Zhang, Shuo Ren, Shujie Liu, Jianyong Wang, Peng Chen, Mu~Li, Ming Zhou,
  and Enhong Chen. 2018.
\newblock Style transfer as unsupervised machine translation.
\newblock \emph{arXiv preprint arXiv:1808.07894}.

\end{thebibliography}
\bibliographystyle{acl_natbib}
% \end{spacing}

\appendix

\section{Ablation Studies}
\label{appendix:ablation}
The following table shows results of the proposed system with ablations on adversarial loss (\textsc{UNTS-adv}) and diversification loss (\textsc{UNTS-div}).
\begin{table}[H]
\centering
\scriptsize
\begin{tabular}{lllll}
\toprule
\textbf{System} & \textbf{FE-diff} & \textbf{SARI} & \textbf{BLEU} & \textbf{Word-diff} \\
\midrule
\textsc{UNTS}+10K & 10.45 & 35.29 & 76.13 & 2.38 \\
\textsc{UNTS-div}+10K & 11.32 & 35.24 & 75.59 & {2.61} \\ %9200
\textsc{UNTS-adv}+10K & 10.32 & 35.08 & 76.19 & {2.64} \\ %10000
\midrule
\textsc{UNTS} & 11.15 & 33.8 & 74.24 & {3.55} \\
\textsc{UNTS-div} & 14.15 & 34.38 & 68.65 & {3.46}\\
\textsc{UNTS-adv} & 12.13 & 34.74 & 73.21 & {2.72}\\\hline
\end{tabular}
\caption{\textsc{UNTS-adv} does not use the adversarial loss, \textsc{UNTS-div} does not use the diversification loss.}
\label{tab:ablation-1} 
\end{table}
\section{Effects of Variation in Labeled Data Size}
\label{appendix:datavar}
The following table shows the effect of labeled data size on the performance of the system. We supplied the system with 2K, 5K, and 10K pairs of complex and simple sentences. From the trained models, models with similar word-diff are chosen for fair comparison. Our observation is that, with increasing data, BLEU as well as SARI increases.
\begin{table}[H]
\centering
\scriptsize
\begin{tabular}{lllll}
\toprule
\textbf{System} & \textbf{FE-diff} & \textbf{SARI} & \textbf{BLEU} & \textbf{Word-diff} \\
\midrule
\textsc{UNTS}+10K & 11.65 & 35.14 & 75.71 & {3.05} \\ %10200
\textsc{UNTS}+5K & 11.69 & 34.39 & 70.96 & {3.01} \\ %10000
\textsc{UNTS}+2K & 11.64 & 34.17 & 72.63 & {3.26} \\
\textsc{UNTS} & 11.15 & 33.8 & 74.24 & {3.55}\\ \hline %11200
\end{tabular}
\caption{Effect of variation in labeled data considered as additional help during training the unsupervised systems.}
\label{tab:ablation-2} 
\end{table}

\end{document}